\documentclass{article}
\pdfoutput=1

    \PassOptionsToPackage{numbers, compress}{natbib}

\usepackage[final]{style} 

\usepackage[utf8]{inputenc} 
\usepackage[T1]{fontenc}    
\usepackage{hyperref}       
\usepackage{url}            
\usepackage{booktabs}       
\usepackage{amsfonts}       
\usepackage{nicefrac}       
\usepackage{microtype}      
\usepackage{xcolor}         
\usepackage{amsmath,bm}
\usepackage{float}
\usepackage{multirow, graphicx}
\usepackage{subfigure}
\usepackage{graphicx}
\usepackage{tabularx, makecell, multirow} 
\usepackage{wrapfig}
\usepackage{amsmath,amssymb}

\newcommand{\ssecspace}{\vspace{-0.4em}}
\newcommand{\secspace}{\vspace{-1.0em}}

\title{SuperVINS: A Real-Time Visual-Inertial SLAM Framework for Challenging Imaging Conditions}

\author{%
  \href{https://luohongkun.com/}{Hongkun Luo}$^\textbf{1}$, Yang Liu$^\textbf{2}$, Chi Guo$^\textbf{3}$\thanks{Corresponding author.} , Zengke Li$^\textbf{4}$, Weiwei Song$^\textbf{5}$
  \\
  \\
  $^1$\href{https://www.zhiyuteam.com/}{Beidou Robots And Intelligent Navigation laboratory}, Wuhan University\\
}

\begin{document}
\bibliographystyle{plain}

\maketitle
\begin{abstract}
 There is an increasing emphasis on achieving high accuracy and robustness in SLAM systems. The traditional visual-inertial SLAM system often struggles with stability under low-light or motion-blur conditions, leading to potential lost of trajectory tracking. High accuracy and robustness are essential for the long-term and stable localization capabilities of SLAM systems. Addressing the challenges of enhancing robustness and accuracy in visual-inertial SLAM, this paper propose SuperVINS, a real-time visual-inertial SLAM framework designed for challenging imaging conditions. In contrast to geometric modeling, deep learning features are capable of fully leveraging the implicit information present in images, which is often not captured by geometric features. Therefore, SuperVINS, developed as an enhancement of VINS-Fusion, integrates the deep learning neural network model SuperPoint for feature point extraction and loop closure detection. At the same time, a deep learning neural network LightGlue model for associating  feature points is integrated in front-end feature matching. A feature matching enhancement strategy based on the RANSAC algorithm is proposed. The system is allowed to set different masks and RANSAC thresholds for various environments, thereby balancing computational cost and localization accuracy. Additionally, it allows for flexible training of specific SuperPoint bag of words tailored for loop closure detection in particular environments. The system enables real-time localization and mapping. Experimental validation on the well-known EuRoC dataset demonstrates that SuperVINS is comparable to other visual-inertial SLAM system in accuracy and robustness across the most challenging sequences. This paper analyzes the advantages of SuperVINS in terms of accuracy, real-time performance, and robustness. To facilitate knowledge exchange within the field, we have made the code for this paper publicly available. You can find the code at this link \url{https://github.com/luohongk/SuperVINS}
\end{abstract}

 \begin{figure}[H]
    \centering
    \includegraphics[width=0.9\linewidth]{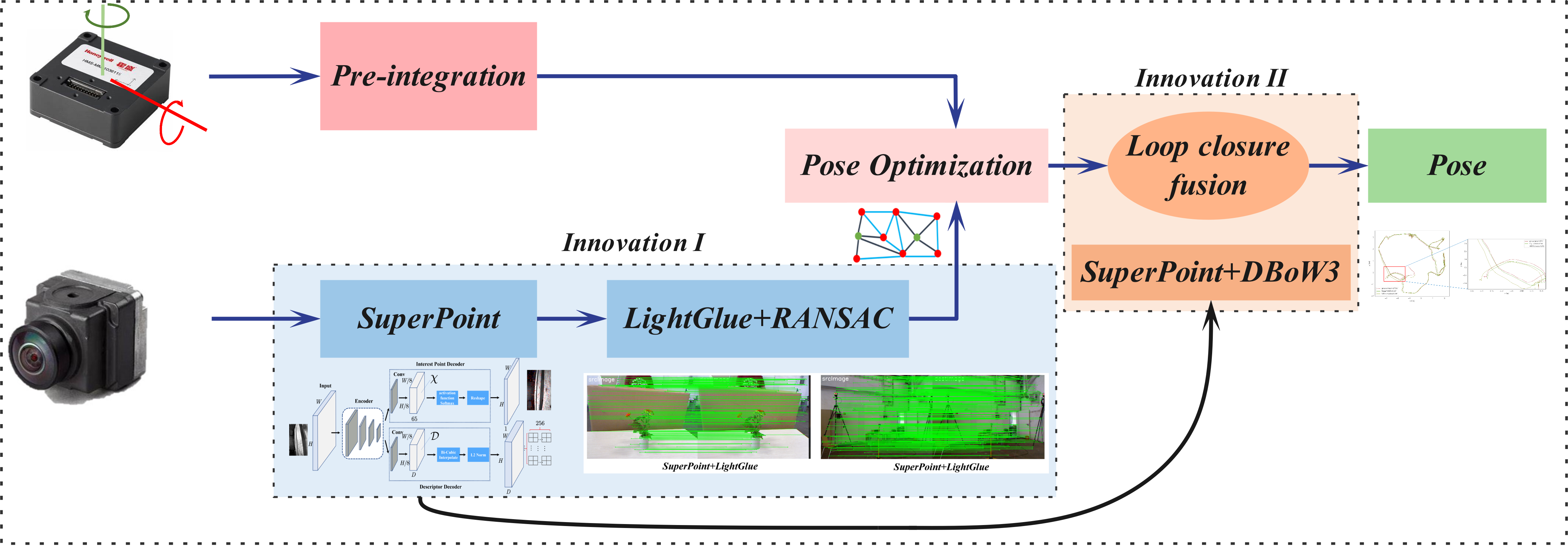}
    \caption{System Introduction}
    \label{SystemIntroduction}
\end{figure}

\secspace
\section{Introduction}
\ssecspace
Autonomous localization and mapping in complex environments is a prominent research direction in the field of robot autonomous intelligent navigation. It not only allows the robot to accurately estimate its own pose, but also continuously senses the surrounding environment. Currently, researchers are considering how to more fully and more effectively utilize the camera and inertial sensor data. Making better use of sensor data can enhance the accuracy and robustness of SLAM systems. A large amount of outstanding SLAM frameworks have already been proposed. These SLAM frameworks are roughly divided into sensor data input, vision front-end feature extraction, input data tracking, back-end pose estimation and optimization, loop closure detection parts.

After extensive exploration and unwavering dedication over the years, many groundbreaking works have emerged in the field of visual inertial SLAM. These works have not only undergone continuous optimization and enhancement by subsequent scholars, but also have exerted a profound influence. The classic visual inertial SLAM framework includes VINS-Mono\cite{vins-mono}, VINS-Fusion\cite{vins-fusion}, ORB-SLAM3\cite{orbslam3}, OpenVINS\cite{OpenVINS}, OKVIS \cite{okvis}, SVO\cite{svo}, MSCKF\cite{msckf}, etc. These well-established visual and visual-inertial SLAM algorithms demonstrate commendable performance in localization and mapping under typical conditions. Nonetheless, this necessitates certain prerequisites, including optimal scene illumination, the absence of motion blur in each frame. When the SLAM system is deployed in real-world scenarios, it often encounters bad environmental conditions. As a result, factors such as inadequate lighting and excessive motion inevitably contribute to the occurrence of blurred image frames. Consequently, the development of a real-time visual-inertial SLAM system capable of addressing these challenges has emerged as a crucial area of research. If this system can not only maintain high localization accuracy but also exhibit strong adaptability and robustness in varying environments, it would prove highly practical, thereby facilitating the broader application of visual-inertial SLAM technology.

To tackle these challenges, we developed SuperVINS, an innovative visual-inertial SLAM framework that leverages deep learning features. On one hand, we incorporated advanced deep learning neural network models in the front-end feature extraction and matching processes of the system. On the other hand, we retrained a bag-of-words model utilizing deeplearning features, which was subsequently employed for loop closure detection within the system. The system demonstrates robust stability and localization accuracy even in conditions of challenging imaging
conditions, we can adaptively train the bag-of-words model based on the specific environmental context. In summary, the contributions of this paper are as follows:

\begin{enumerate}{}{}
\item{
This paper integrates deep learning feature extraction and matching methods into the VINS-Fusion front end and proposes a matching enhancement strategy based on RANSAC algorithm. The powerful feature extraction and matching capabilities of deep learning can yield superior and more robust features for back-end optimization, thereby enhancing the overall performance of the SLAM system.
}
\item{This paper introduces the integration of deep learning features for loop closure detection within the VINS-Fusion framework. It implements the training and integration of deep learning bag-of-words models tailored to specific datasets, thereby ensuring reliable localization while preserving real-time performance.}

\item{
The real-time visual-inertial SLAM code, which incorporates deep learning features, has been open-sourced at \url{https://github.com/luohongk/SuperVINS}. This initiative is expected to facilitate technical exchanges among researchers in the SLAM community.
}

\end{enumerate}

\secspace
\section{Related Work}
This section provides a review of classical visual-inertial SLAM methods, primarily focusing on filtering and graph optimization techniques. Additionally, it highlights research efforts related to deep learning-based feature matching methods and the integration of deep learning approaches with visual-inertial SLAM.

\subsection{Traditional visual-inertial SLAM}
In recent years, substantial research has been conducted in the field of visual-inertial SLAM, resulting in the development of several well-established systems based on traditional visual geometric features. SLAM solutions can be broadly categorized into two primary approaches: filtering-based methods and graph optimization-based methods.

MSCKF \cite{msckf} represents an early approach to visual-inertial odometry utilizing a direct method, employing an extended Kalman filter for state estimation. In 2020, OpenVINS \cite{OpenVINS} was introduced as a visual-inertial odometry system based on geometric features, utilizing a manifold sliding window Kalman filter for state estimation. In 2023, Giulio Delama et al. introduced UVIO\cite{uvio}, which incorporated Ultra-Wideband (UWB) technology into the sensor framework and employed Kalman filtering to address pose estimation. Additionally, in the same year, researchers implemented equivariant filtering in visual-inertial odometry\cite{eqvio}, developing a novel Lie group symmetry to enhance the filtering consistency of the VIO problem. This framework is referred to as EqVIO.

Notably, the most prominent system in the realm of graph optimization-based methods is VINS-Mono\cite{vins-mono}, developed by Tong Qin et al. in 2018. VINS-Mono is a tightly coupled visual-inertial SLAM framework, wherein the visual front end eliminates the need for feature point matching, thereby enhancing operational efficiency. Building upon the VINS-Mono framework, researchers subsequently integrated GPS sensors to introduce VINS-Fusion\cite{vins-fusion}. In 2021, Carlos Campos et al. enhanced the capabilities of ORB-SLAM2\cite{orbslam2}, a purely visual SLAM system, by incorporating inertial sensors, resulting in the development of ORB-SLAM3\cite{orbslam3}—a widely recognized open-source visual-inertial SLAM framework grounded in graph optimization. In 2024, Nathaniel Merrill et al. introduced AB-VINS\cite{ab-vins}, a visual-inertial SLAM system that forgoes the estimation of sparse feature positions. Instead, it focuses exclusively on estimating the scale and bias parameters of the monocular depth map while employing an innovative memory tree structure. This framework exemplifies a sophisticated approach to visual-inertial SLAM grounded in graph optimization.

Regardless of whether a SLAM system is based on filtering or graph optimization, the image feature extraction module plays a crucial role.

\subsection{Feature Extraction and Matching Method Based on Deep Learning}

In the past few years, numerous solutions for feature point extraction and matching using deep learning have emerged. However, this paper aims to integrate these methods into the SLAM system, with a primary focus on significant open-source contributions related to feature point extraction and matching.

In 2018, Daniel DeTone et al. introduced SuperPoint\cite{superpoint}, a self-supervised neural network for feature point extraction that delivers high-quality feature points and generates fixed-length feature descriptors. This method is regarded as a classic in the field and serves as a primary approach utilized in this paper. In 2020, researchers introduced the DISK\cite{disk} end-to-end feature extraction method, which employs reinforcement learning to extract high-density feature points. However, in the context of SLAM systems, an excessive density of feature points can adversely impact the speed of feature matching. Subsequently, with the advent of powerful architectures such as transformers, Jiaming Sun et al. introduced Loftr\cite{loftr} in 2020. While this method addressed the challenges of image feature extraction in low-texture scenarios to some extent, the incorporation of complex transformers led to significant computational resource demands, which hindered its ability to meet real-time performance requirements. In 2024, Guilherme Potje et al. introduced XFeat\cite{xfeat}, a feature extraction and matching method that addresses the operational speed issues associated with Loftr. By leveraging a CNN architecture, XFeat is designed for efficient operation on low-cost CPUs. However, the 64-dimensional floating-point vectors generated by XFeat result in suboptimal performance for loop closure detection due to their relatively low dimensionality, as their relatively low dimensionality limits their effectiveness in this context.

There has been significant progress in deep learning-based feature matching. In 2020, Paul-Edouard Sarlin et al. introduced the SuperGlue\cite{superglue} method, which exhibits robust feature matching capabilities. However, it still faces challenges in low-texture scenarios, indicating that there is potential for further improvement. Subsequently, researchers conducted extensive studies on feature matching. In 2023, Philipp Lindenberger et al. introduced LightGlue\cite{lightglue}, a lightweight, efficient, and high-accuracy feature extraction method. Its lightweight design makes it particularly well-suited for integration into visual-inertial SLAM systems. In 2024, researchers introduced XFeat, a feature matching method that integrates multiple matching modes. However, it remains constrained by the dimensionality of its descriptors, which limits its effectiveness under challenging imaging conditions. In the same year, Hanwen Jiang et al. introduced OmniGlue\cite{omniglue}, a method designed to enhance the generalization capabilities of feature matching networks across diverse scenarios, without prioritizing real-time system deployment. However, the lengthy matching times associated with OmniGlue render it unsuitable for SLAM systems, where frequent feature matching is essential. Recently, researchers including Yifan Wang have introduced Efficient LoFTR\cite{efficientloftr}, an advancement built upon the original LoFTR framework. Compared to the 2021 version, Efficient LoFTR simplifies the model architecture and achieves enhanced sub-pixel matching accuracy. While this represents a significant improvement, it remains reliant on the Transformer architecture, which may lead to performance bottlenecks when deployed on resource-constrained devices.

In recent times, there has been a continuous evolution of deep learning-based feature extraction and matching methods. However, for the practical needs of visual-inertial SLAM, high-dimensional feature descriptors are essential for more accurate loop closure detection. Additionally, real-time SLAM systems necessitate lightweight feature matching methods to satisfy their real-time requirements. Given these comprehensive considerations, the use of SuperPoint and LightGlue is particularly suitable in this context.

\subsection{Visual-inertial SLAM based on deep learning}
Visual-inertial SLAM leveraging deep learning can be categorized into two primary approaches. Firstly, numerous researchers employ deep learning techniques to replace specific modules within the SLAM system. Secondly, some researchers adopt end-to-end deep learning methods to directly estimate the camera pose.

In 2023, the VINS-FEN\cite{vinsfen} framework was proposed, it incorporates the CNN-based feature extraction network FEN into the VINS architecture, yielding promising results.However, it does not implement a feature matching network and relies on geometric distances of the descriptors for matching, necessitating improvements in matching performance. Additionally, this system requires offline preprocessing for feature extraction, thus limiting its capability for real-time processing. In 2024, D-VINS \cite{dvins} was introduced as a visual-inertial SLAM method that employs deep learning for feature extraction and matching exclusively in loop closure detection. It does not fully integrate deep learning features into the VINS-Fusion front end and loop closure detection, indicating significant potential for improvement. All of these approaches leverage deep learning solutions to enhance specific components of the SLAM system. 
In 2017, Sen Wang et al. introduced DeepVIO\cite{deepvio}, a self-supervised end-to-end training network for monocular visual-inertial odometry (VIO). Subsequently, in 2023, researchers proposed an end-to-end neural network approach integrated with a Kalman filter\cite{endtoendvio}. However, it is important to note that end-to-end deep learning methods often lack a certain degree of interpretability.

\section{System Overview}
\begin{figure}[h]
    \centering
    \includegraphics[width=0.9\linewidth]{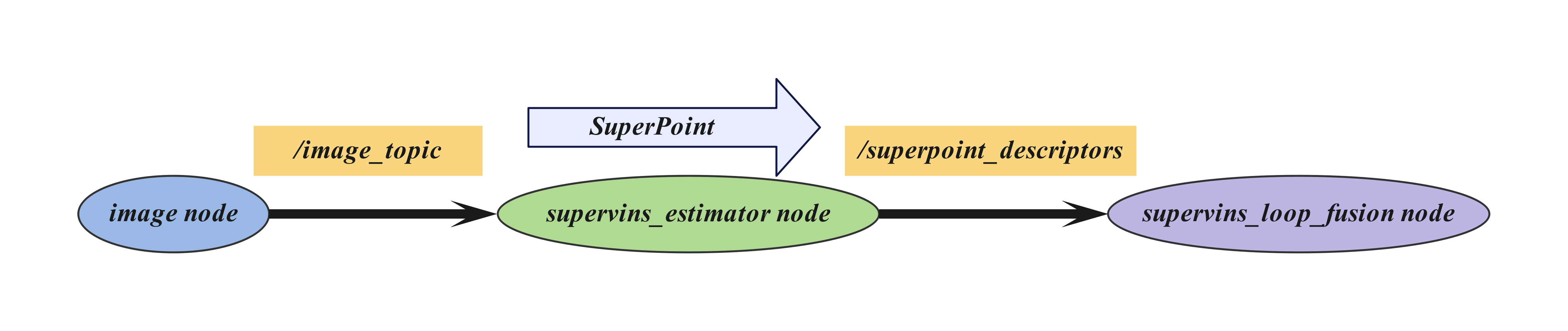}
    \caption{Schematic of deep learning descriptor transfer.}
    \label{node}
\end{figure}

\begin{figure*}[htp] 
  \includegraphics[width=1\textwidth]{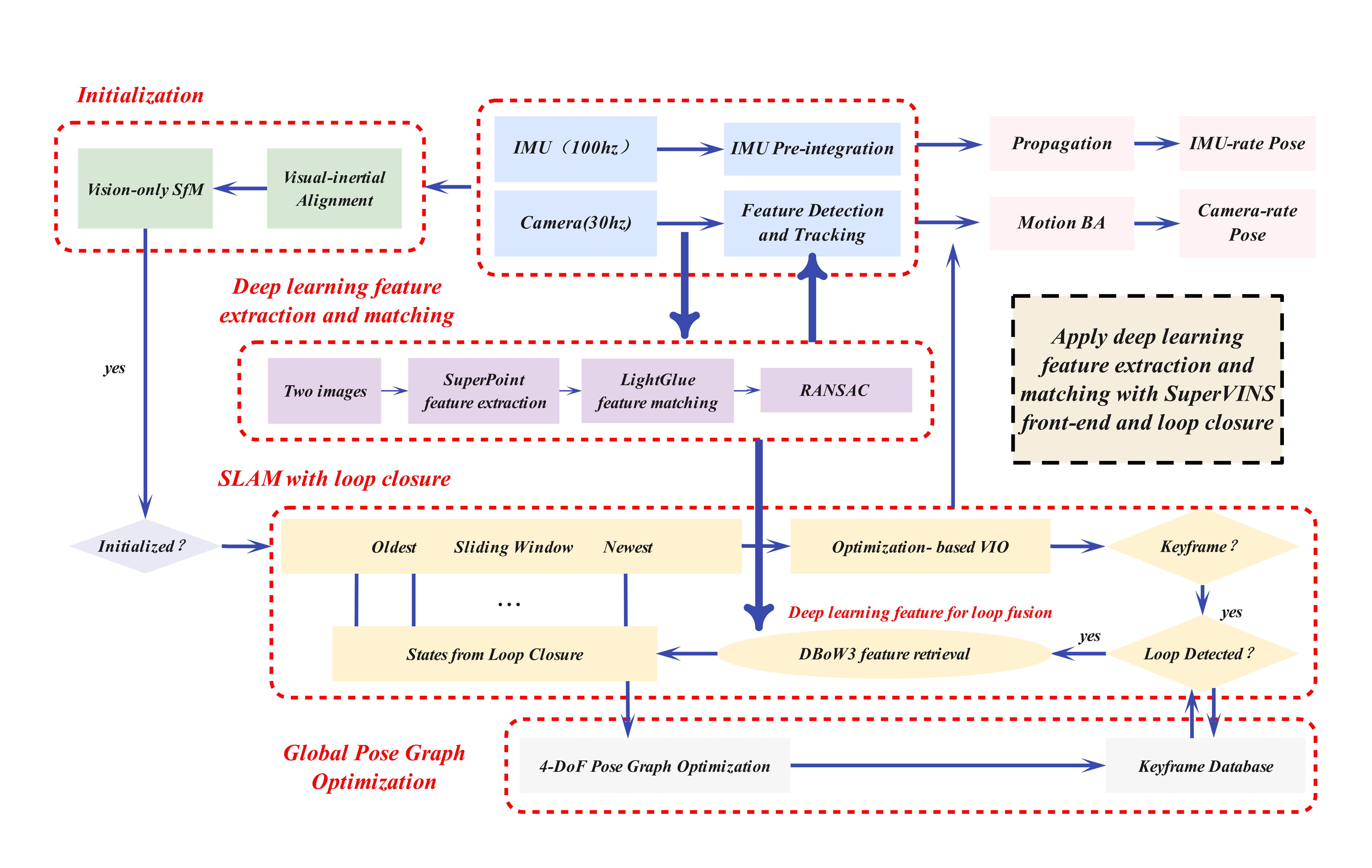}
  \caption{Overview of the SuperVINS framework: The system integrates camera and IMU data as input. It employs SuperPoint and LightGlue to match features between consecutive image frames while performing pre-integration. After the LightGlue matching process, SuperVINS utilizes the RANSAC algorithm to enhance the accuracy of feature correspondences. Upon completion of front-end optimization, the extracted features are synchronously transmitted to the node responsible for loop closure detection. SuperVINS constructs keyframes and conducts sliding window-based pose optimization. After pose calculations, the system relays the features, pose, and point cloud maps of the keyframes to the loop closure detection node, which employs DBoW3 for feature retrieval and pose graph optimization.}
  \label{SuperVINSOverview}
\end{figure*}

This system utilizes a visual-inertial measurement framework comprising a monocular camera and an IMU. It builds upon VINS-Fusion for enhancement and integration. As illustrated in Fig. \ref{SuperVINSOverview}, this paper elaborates on four key components of the methodology: the feature extraction module, feature matching module, feature matching optimization module, and loop closure detection module. The SuperVINS system is implemented within the ROS framework for data communication, with the deep learning descriptor ROS node and data transfer depicted in Fig. \ref{node}. In the back-end, we adhere to the sliding-window-based pose optimization strategy employed by VINS-Fusion.

\section{Method}
\label{headings}

This section is organized into three key components. The first component provides a concise overview of the deep learning-based feature extraction and matching techniques employed by SuperVINS , This part will be introduced in \ref{feature_sp_lg}. The second component discusses the matching enhancement strategy, which utilizes the RANSAC algorithm. This part will be introduced in \ref{RANSAC_OP}. Lastly, the third component focuses on the loop detection mechanism that leverages a bag-of-words approach. This part will be introduced in \ref{loop_FU}.

\subsection{SuperPoint Feature Extraction and LightGlue Feature Matching}\label{feature_sp_lg}

\begin{figure}[h]
    \centering
    \includegraphics[width=0.9\linewidth]{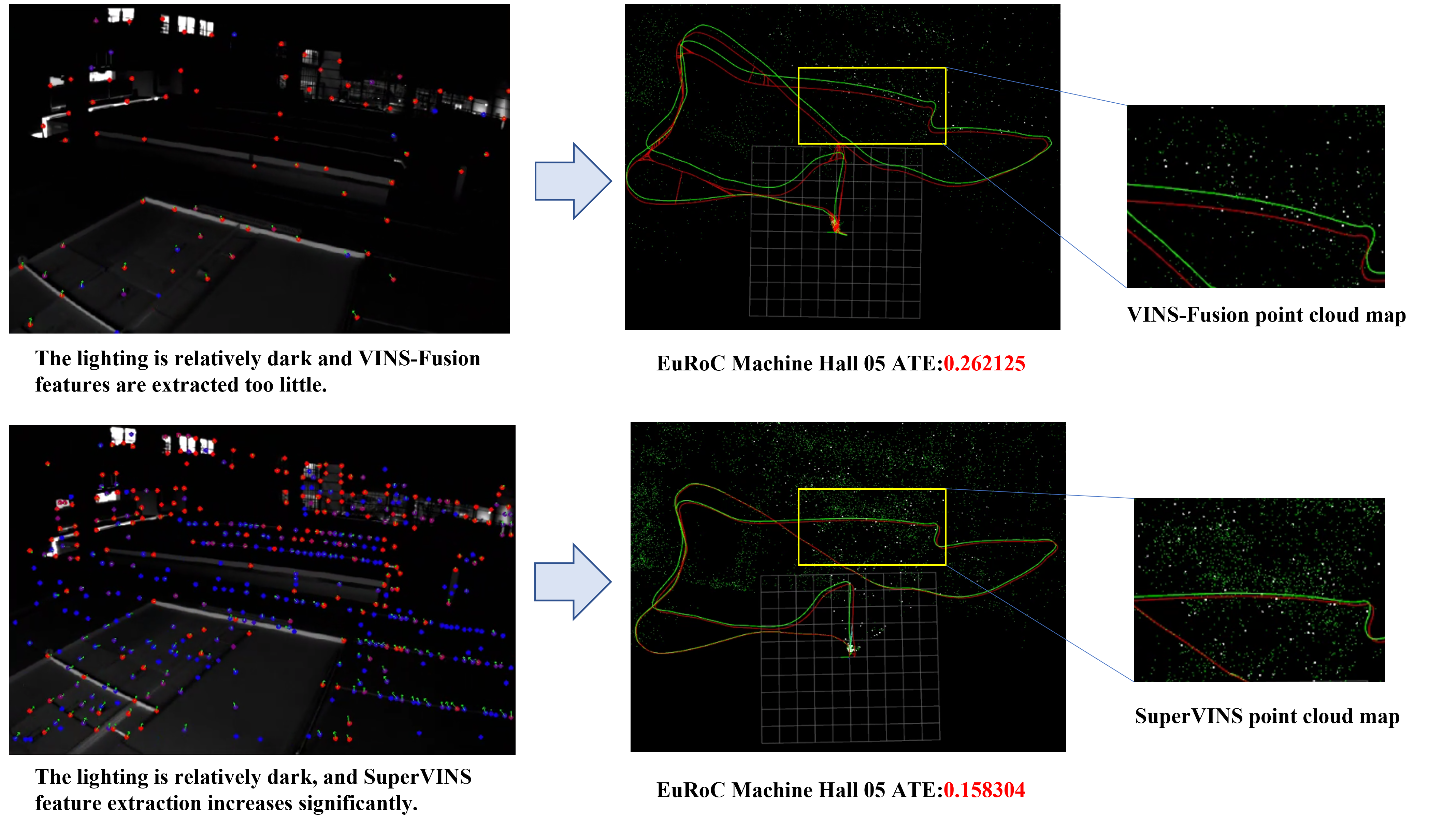}
    \caption{Comparison of Traditional Geometric Feature Points and Deep Learning Feature Points in Visual Inertial SLAM.}
    \label{compare_map}
\end{figure}

In the feature extraction component, this study adopts SuperPoint\cite{superpoint} deep learning features. The SuperPoint network provides two significant advantages. At first, it not only identifies a robust number of feature points but also generates a fixed-length descriptor of 256 dimensions. Besides, This dual capability enhances the effectiveness and reliability of feature representation, making it well-suited for diverse applications in SLAM system. The swift movement of the camera and inadequate lighting  can lead to suboptimal image quality. In such scenarios, insufficient extraction of feature points and inadequate overlap between consecutive frames can significantly increase the risk of tracking loss. Consequently, the robust feature extraction capabilities of SuperPoint effectively address these challenges, ensuring reliable tracking even under adverse conditions. fixed-length descriptors generated by SuperPoint is advantageous for training deep learning bag-of-words models. This is particularly important.Because the dimensionality of the input $N\times dim$ matrix must remain consistent throughout the bag-of-words training and application process. Where $N$ represents the number of feature points in an image. Therefore, SuperPoint features are exceptionally well-suited for visual-inertial SLAM systems, offering a reliable foundation for effective feature representation and processing. Fig. \ref{compare_map} presents a result of classical geometric features and deep learning-based features within the framework of SLAM, with a particular focus on their performance in scenarios characterized by poor lighting conditions.

\begin{figure}[h]
    \centering
    \includegraphics[width=0.9\linewidth]{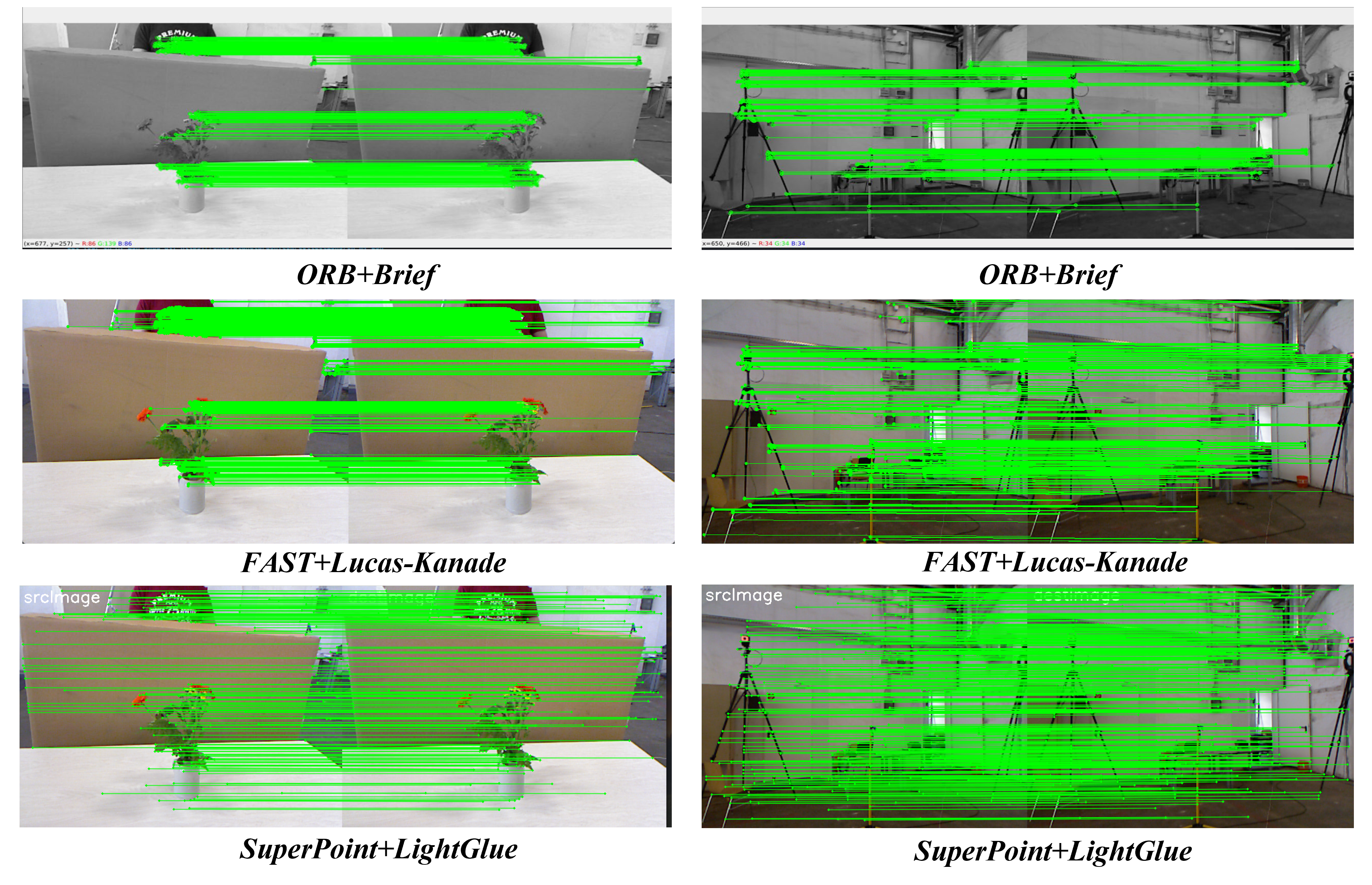}
    \caption{Matching Effect Comparison.The figure illustrates three methods of feature extraction and association for the same pair of images. It is evident that utilizing deep learning for feature extraction and matching yields superior results.}
    \label{match}
\end{figure}

LightGlue \cite{lightglue} is a highly efficient feature extraction network. Operational speed of feature matching directly influencing the overall efficiency of the SLAM system. Lightweight architecture of LightGlue and ease of deployment are key factors that make this feature matching network an ideal choice for the present study. Moreover, this feature matching network is capable of delivering robust performance in both indoor and outdoor environments. Notably, in scenarios characterized by blurred images and low lighting, it can dynamically adjust its operational rate in response to the complexity of the matching task. This adaptive capability further enhances the real-time performance of the system, ensuring reliable operation under varying conditions. This paper aims to compare the LightGlue feature matching approach with traditional geometric feature-based methods, thereby highlighting the advantages of the LightGlue scheme. The results of this comparison are illustrated in Fig. \ref{match}.

\subsection{Matching Enhancement Strategy}\label{RANSAC_OP}
In comparison to traditional visual-inertial odometry, deeplearning visual features employed for extraction and matching tend to generate a larger number of matched point pairs. However, this increase also raises the likelihood of incorrectly matched pairs. This paper seeks to improve the extraction of matched feature point pairs by reducing the radius of the masking region. The mask serves to prevent feature points from clustering too closely, thereby facilitating the identification of additional matched feature point pairs within the specified radius.By reducing the radius of the mask, the number of point pairs increases, resulting in a denser distribution that offers more candidates for subsequent RANSAC optimization. The threshold for the RANSAC algorithm is set to a very low value, imposing stringent criteria for correct matching and thereby ensuring the quality of accurately matched point pairs.

During the RANSAC optimization of feature matching, four pairs of feature points are randomly selected from the matching results.For each pair of feature points,there exists a corresponding matching point counterpart $\boldsymbol{x}_2=H\boldsymbol{x}_1$ on the normalization plane, where $\boldsymbol{x}_1$ and $\boldsymbol{x}_2$ denote the matching point counterparts,and represents the homography matrix. The homography matrix $H$ has 8 degrees of freedom and exhibits scale invariance. For each pair of matching points,as expressed in Equation \ref{ransac1}:

\begin{equation}
\left. \left[ \begin{array}{c}
	u_2\\
	v_2\\
	1\\
\end{array} \right. \right] =\left[ \begin{matrix}
	h_1&		h_2&		h_3\\
	h_4&		h_5&		h_6\\
	h_7&		h_8&		h_9\\
\end{matrix} \right] \left[ \begin{array}{c}
	u_1\\
	v_1\\
	1\\
\end{array} \right] 
\label{ransac1}
\end{equation}

If the third constraint is substituted, the following conclusion can be obtained as in Equation \ref{ransac2}:

\begin{equation}
 \begin{aligned}
	u_2&=\frac{h_1u_1+h_2v_1+h_3}{h_7u_1+h_8v_1+h_9}\\
	v_2&=\frac{h_4u_1+h_5v_1+h_6}{h_7u_1+h_8v_1+h_9}\\
\end{aligned}
\label{ransac2}
\end{equation}

After organizing the above equations, we can get the following Equation \ref{ransac3}

\begin{equation}
\left\{ \begin{array}{l}
	h_1u_1+h_2v_1+h_3-h_7u_1u_2-h_8u_1u_2-h_9v_2=0\\
	h_4u_1+h_5v_1+h_6-h_7u_1u_2-h_8u_1u_2-h_9v_2=0\\
\end{array} \right. 
\label{ransac3}
\end{equation}

Each point correspondence generates two constraint equations, allowing a homography matrix with 8 degrees of freedom to be determined using only 4 pairs of point correspondences. To address this, we formulate a least squares problem. Given the scale invariance, we can directly impose the condition $h_9=1$.

The set of feature points to be extracted for each image frame is $N=\left\{ p_{N}^{1},p_{N}^{2},p_{N}^{3}\cdots p_{N}^{t} \right\} $ and t is the number of feature points in the image. Therefore, for image A, the set of feature points is $N_A=\left\{ p_{N_A}^{1},p_{N_A}^{2},p_{N_A}^{3}\cdots p_{N_A}^{t} \right\} $, and for image B, the set of feature points is $N_B=\left\{ p_{N_B}^{1},p_{N_B}^{2},p_{N_B}^{3}\cdots p_{N_B}^{t} \right\} $. In order to prevent the feature points from aggregating too much, and try to make the feature points uniformly distributed in the image, we are given the radius of the mask $r$. For the points in the set of feature points, we need to make sure that any two feature points, $p_1\left( x_1,y_1 \right) $ and $p_2\left( x_2,y_2 \right) $, satisfy the conditions of Equation \ref{dis}.

\begin{equation}
\left\{ \begin{array}{l}
	dis=\sqrt{\left( x_1-x_2 \right) ^2+\left( y_1-y_2 \right) ^2}\\
	dis<r\\
\end{array} \right. 
\label{dis}
\end{equation}

Building on this foundation, this paper further refines feature point matching using the RANSAC algorithm. The matching performance is enhanced by lowering the inlier threshold. The set of feature points following the aforementioned mask optimization is $N_{A}^{'}$, $N_{B}^{'}$.

\begin{equation}
f\left( N_{A}^{'},N_{B}^{'} \right) <\delta 
\label{FNANB}
\end{equation}

where $f$ is the RANSAC algorithm performed each time and $\delta $ is the reduced interior point threshold. The lower the inner point threshold, the more accurate the feature point matching. After repeating this process many times, it is possible to carry out one step to optimize the feature points after matching. In turn, a sufficient number of feature points with high robustness are obtained. In order to illustrate the advantages of the matching enhancement strategy this paper is followed by a detailed comparison.

where $f$ represents the RANSAC algorithm applied iteratively, while $\delta $ denotes the reduced inlier threshold. A lower inlier threshold results in more accurate feature point matching. After repeating this process multiple times, a subsequent optimization step can be performed on the matched feature points, yielding a sufficient number of highly robust feature points. To demonstrate the advantages of this enhancement strategy, a detailed comparison is presented in the following sections.

\begin{figure}[h]
    \centering
    \includegraphics[width=0.9\linewidth]{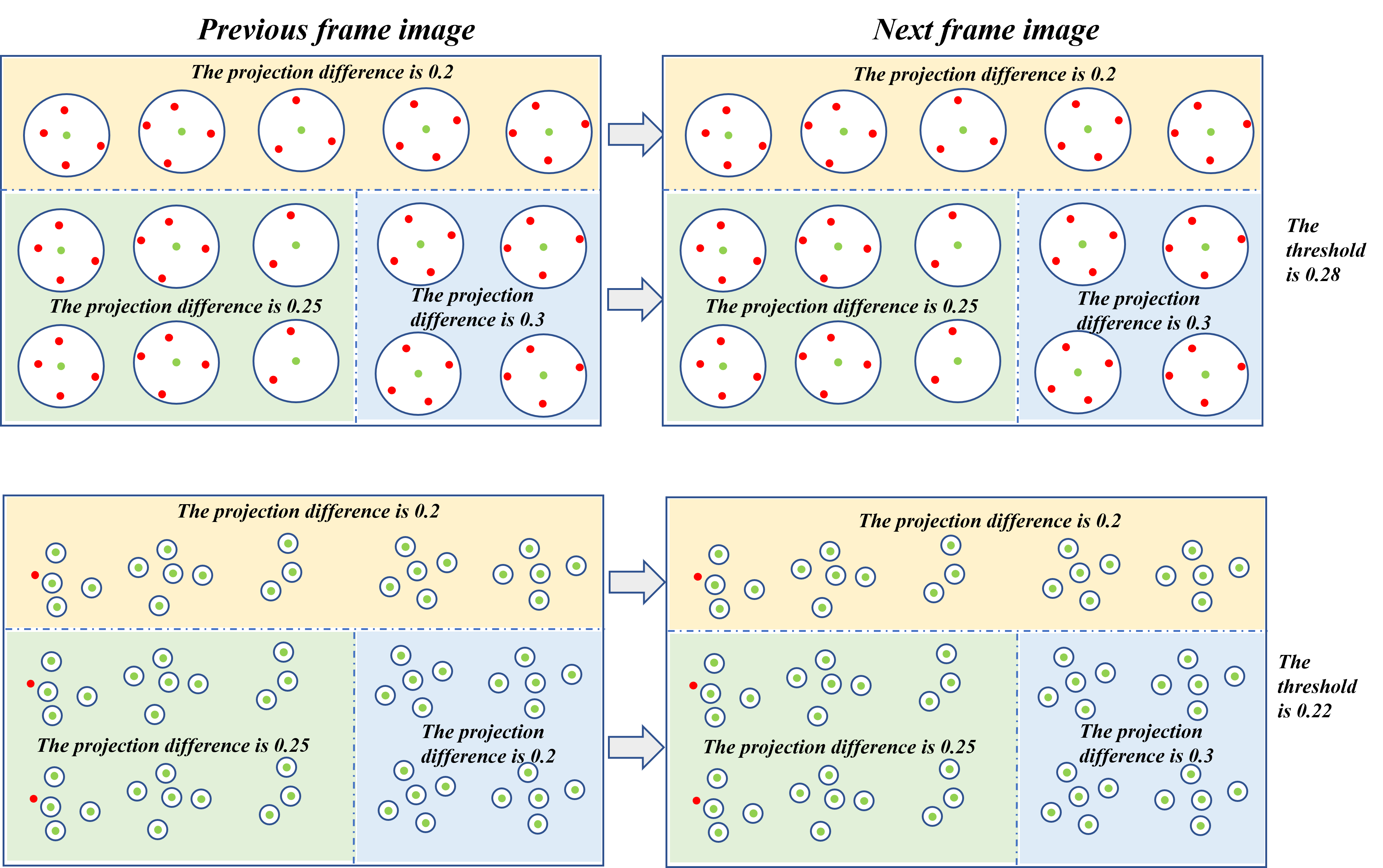}
    \caption{Schematic diagram of feature matching enhancement strategy.}
    \label{match_opti}
\end{figure}

As illustrated in Fig.\ref{match_opti}, the errors from the previous frame are projected onto the next frame using the homography matrix, resulting in projection differences of 0.2, 0.25, and 0.3, respectively. However, due to the large mask radius, many points must be eliminated. When the projection matching threshold is set to 0.28, a total of 11 points are identified that meet the criteria, highlighted in green, with a projection difference of less than 0.28. By reducing the mask radius, more feature points can be retained. As shown in the subsequent plots in Fig.\ref{match_opti}, the number of green feature points increases significantly. When the threshold is further lowered to 0.22, matching accuracy improves, as only points with a projection difference less than 0.22 are considered. At this threshold, 22 points qualify as green points with a projection difference below 0.22. This optimization effectively ensures that a sufficient number of points are extracted while minimizing the projection difference between matched pairs.
\subsection{Loop Closure Detection with Deep Learning}\label{loop_FU}
The Bag of Words (BoW) model was initially developed for text categorization and has since been increasingly applied in location recognition tasks. In this context, multiple images can be represented as a visual Bag of Words by utilizing image features. This approach involves creating a visual dictionary structured as a hierarchical tree, where similar image features are clustered together. To determine which historical image frame is most similar to the current frame, a Bag of Words vector can be generated from the visual dictionary, allowing for efficient searching based on this vector.

The Bag-of-Words vectors employed in well-known visual-inertial SLAM systems such as VINS-Fusion and ORB-SLAM3 are utilized for loop closure detection. However, DBoW2 is relatively slow in generating lexicons for a given dataset. Moreover, generating Bag-of-Words vectors from the descriptors of a single image does not match the speed of DBoW3 when the SLAM system operates in real time. Therefore, in this paper, we create new lexicons specifically tailored to the dataset used in this experiment. The lexicon implemented in SuperVINS is based on the advanced SuperPoint deep learning features, and Bag-of-Words vectors are generated using the more efficient DBoW3.

\section{Experiments}
\label{4}

This experiment is conducted on the Ubuntu 18.04 operating system, utilizing a GeForce RTX 2060 graphics card with NVIDIA driver version 530.41.03, and ROS Melodic. Additionally, the system is configured with CUDA 11.7, CUDNN 8.9.6, and ONNX Runtime 1.16.3 to enable inference of deep learning models. To demonstrate the advantages of the improved algorithm, we employ the well-known public EuRoC dataset for our experiments. The system uses a re-trained deep learning-based Bag-of-Words model. To ensure comprehensive coverage of various scenarios, we expand the training dataset to include a hybrid set comprising the EuRoC, TUM, and KITTI datasets.

\subsection{Accuracy Testing of SuperVINS}
This section evaluates the localization accuracy of the enhanced algorithm. The accuracy metrics used are the root mean square error of the Absolute Trajectory Error (ATE), the root mean square error of the Relative Position Error (RPE) in translation (RPEt), and the root mean square error of the Relative Rotation Error (RPEr).

\begin{figure}[H]
    \centering
    \includegraphics[width=0.9\linewidth]{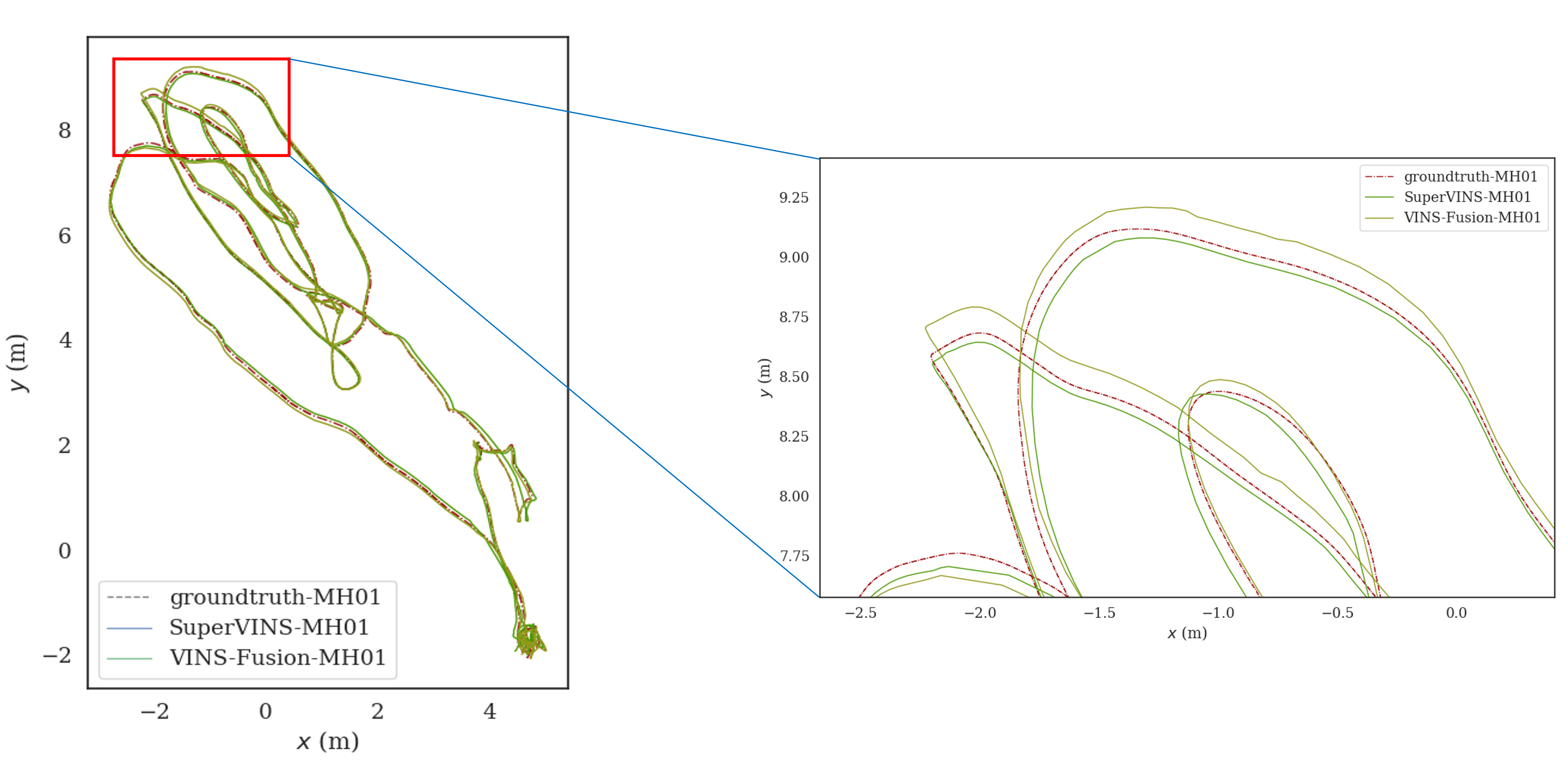}
    \caption{Trajectory comparison chart on the MH01 sequence.}
    \label{guiji1}
\end{figure}

\begin{figure}[H]
    \centering
    \includegraphics[width=0.9\linewidth]{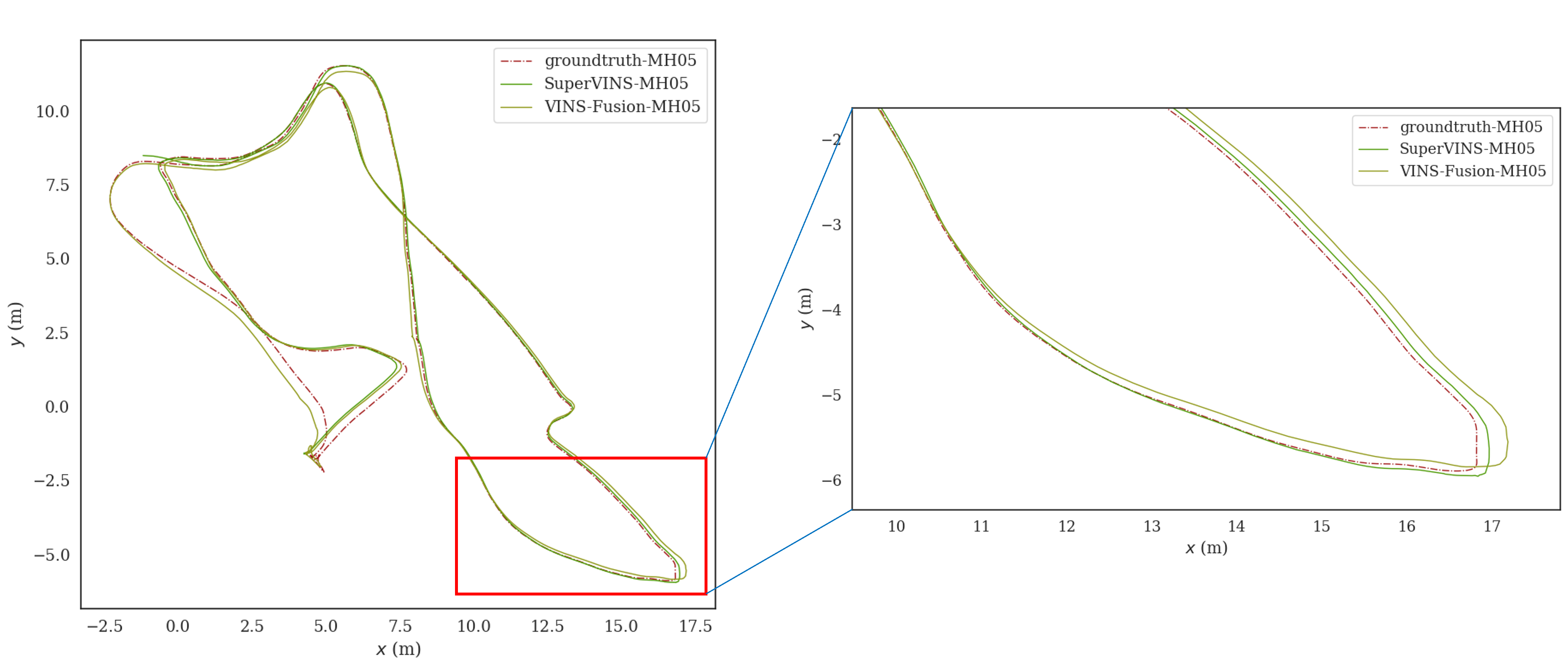}
    \caption{Trajectory comparison chart on the MH05 sequence.}
    \label{guiji2}
\end{figure}

\begin{figure}[H]
    \centering
    \includegraphics[width=0.9\linewidth]{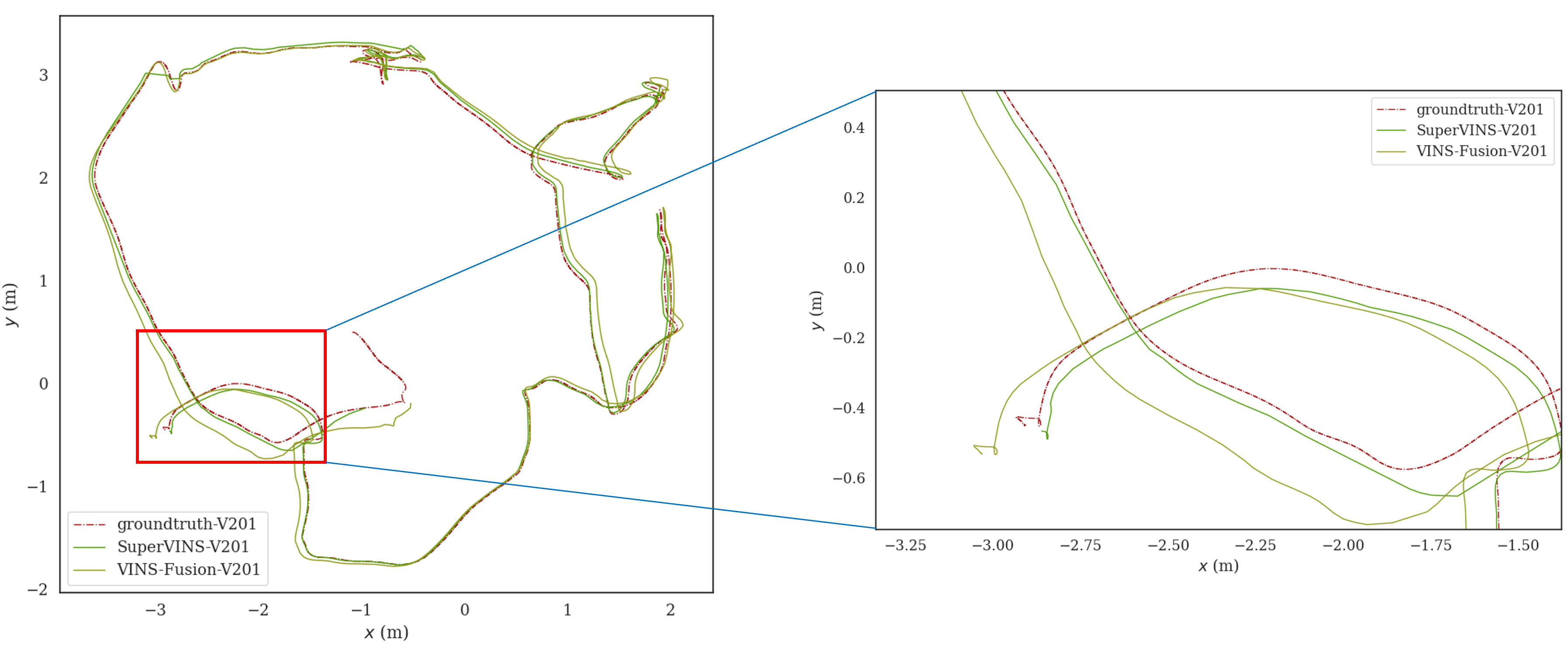}
    \caption{Trajectory comparison chart on the V201 sequence.}
    \label{guiji3}
\end{figure}

To highlight the effectiveness of the improved algorithm, this section conducts experiments on 11 sequences from the EuRoC dataset. The experiments are organized into two primary components. First, SuperVINS is compared with the baseline algorithm VINS-Fusion. Second, SuperVINS is assessed against various algorithms in terms of Absolute Trajectory Error (ATE) within the monocular + IMU mode. Additionally, we perform statistical analyses and trajectory visualizations of the trajectory errors. The results of the trajectory visualizations are presented in Fig. \ref{guiji1}, \ref{guiji2}, \ref{guiji3}, effectively illustrating the advantages of SuperVINS.

\begin{table*}[]
  \caption{THE ATE OF VINS-Fusion(MONO+IMU) AND SuperVINS(MOno+IMU) ON THE EuRoC DATA SET, THE RMSE OF RPEr AND RPEt, AND THE BETTER ATE OF SuperVINS ARE BOLDED}
  \centering
  \label{vins-fusion_SuperVINS_compare}
\begin{tabular}{l|ccc|ccc}
\toprule
  & \multicolumn{3}{c|}{VINS-Fusion} & \multicolumn{3}{c}{SuperVINS} \\
\hline
dataset & ATE & RPEr & RPEt & ATE & RPEr & RPEt \\
\midrule
MH01 & 0.091079 & 0.001536 & \textbf{0.006640} & \textbf{0.086656} &\textbf{ 0.002031} & 0.006729 \\
MH02 & \textbf{0.045720} & \textbf{0.001264} & 0.005192 & 0.096954 & 0.002177 & \textbf{0.004465} \\
MH03 & \textbf{0.091861} & 0.005200 & 0.009599 & 0.180578 & \textbf{0.001697} & \textbf{0.00818} \\
MH04 & \textbf{0.130696} & 0.002976 & 0.009822 & 0.170761 & \textbf{0.00199} & \textbf{0.009613} \\
MH05 & 0.262125 & \textbf{0.003364} & 0.013499 & \textbf{0.158304} & 0.003786 & \textbf{0.009489} \\
V101 & \textbf{0.153585} & 0.004278 & \textbf{0.007410} & 0.201021 & \textbf{0.002061} & 0.007983 \\
V102 & 0.219169 & 0.004236 & \textbf{0.008229} & \textbf{0.11599} & \textbf{0.002722} & 0.020395 \\
V103 & \textbf{0.169991} & \textbf{0.004300} & \textbf{0.007352} & 0.22141 & 0.021587 & 0.030906 \\
V201 & 0.123334 & \textbf{0.002993} & \textbf{0.004906} & \textbf{0.061106} & 0.004127 & 0.004992 \\
V202 & 0.000000 & 0.000000 & 0.000000 & \textbf{0.100303} & \textbf{0.00594} &\textbf{ 0.0085} \\
V203 & 0.192608 & \textbf{0.008283} & \textbf{0.013867} & \textbf{0.168746} & 0.017337 & 0.028083 \\
\bottomrule
\end{tabular}
\end{table*}

\begin{table*}[]
  \caption{COMPARISION OF ATE RESULTS OF SIX ALGORITHMS(MONO+IMU)}
  \centering
  \label{six_compare}
 \begin{tabular}{p{1cm} p{1cm} p{1cm} p{1cm} p{1cm} p{1cm} p{1cm}}
\toprule
dataset & OKVIS & VINS-Mono-Noloop & VINS-Mono-loop & VINS-Fusion-Noloop & VINS-Fusion-loop & SuperVINS \\
\midrule
MH01 & 0.3300 & 0.1556 & 0.0868 & 0.1377 & 0.0911 & \textbf{0.0867} \\
MH02 & 0.3700 & 0.1780 & 0.1083 & 0.1002 & 0.0457 & 0.0970 \\
MH03 & 0.2500 & 0.1950 & 0.0658 & 0.1203 & 0.0919 & 0.1806 \\
MH04 & 0.2700 & 0.3470 & 0.2047 & 0.2836 & 0.1307 & 0.1708 \\
MH05 & 0.3900 & 0.3020 & 0.1421 & 0.1587 & 0.2621 & \textbf{0.1583} \\
V101 & 0.0940 & 0.0889 & 0.0484 & 0.1569 & 0.1536 & 0.2010 \\
V102 & 0.1400 & 0.1105 & 0.0631 & 0.1222 & 0.2192 & \textbf{0.1160} \\
V103 & 0.2100 & 0.1875 & 0.2012 & 0.1287 & 0.1700 & 0.2214 \\
V201 & 0.0900 & 0.0863 & 0.0672 & 0.1207 & 0.1233 & \textbf{0.0611} \\
V202 & 0.1700 & 0.1580 & 0.1588 & 0.0000 & 0.0000 & \textbf{0.1003} \\
V203 & 0.2300 & 0.2775 & 0.2475 & 0.2881 & 0.1926 & \textbf{0.1687} \\
\bottomrule
\end{tabular}
\end{table*}

In comparing SuperVINS with classic algorithms such as OKVIS, VINS-Mono-Noloop, VINS-mono-loop, VINS-Fusion-Noloop, and VINS-Fusion-loop regarding ATE, we ensured that the algorithms were consistent at the hardware level. As shown in TABLE \ref{six_compare}, SuperVINS demonstrated advantages in 6 out of the 11 challenging scene sequences from the EuRoC dataset. This indicates that the integration of deep learning neural networks, specifically SuperPoint and LightGlue, effectively enhances the localization capabilities of visual-inertial SLAM in extreme conditions.

\begin{figure}[H]
    \centering
    \includegraphics[width=0.5\linewidth]{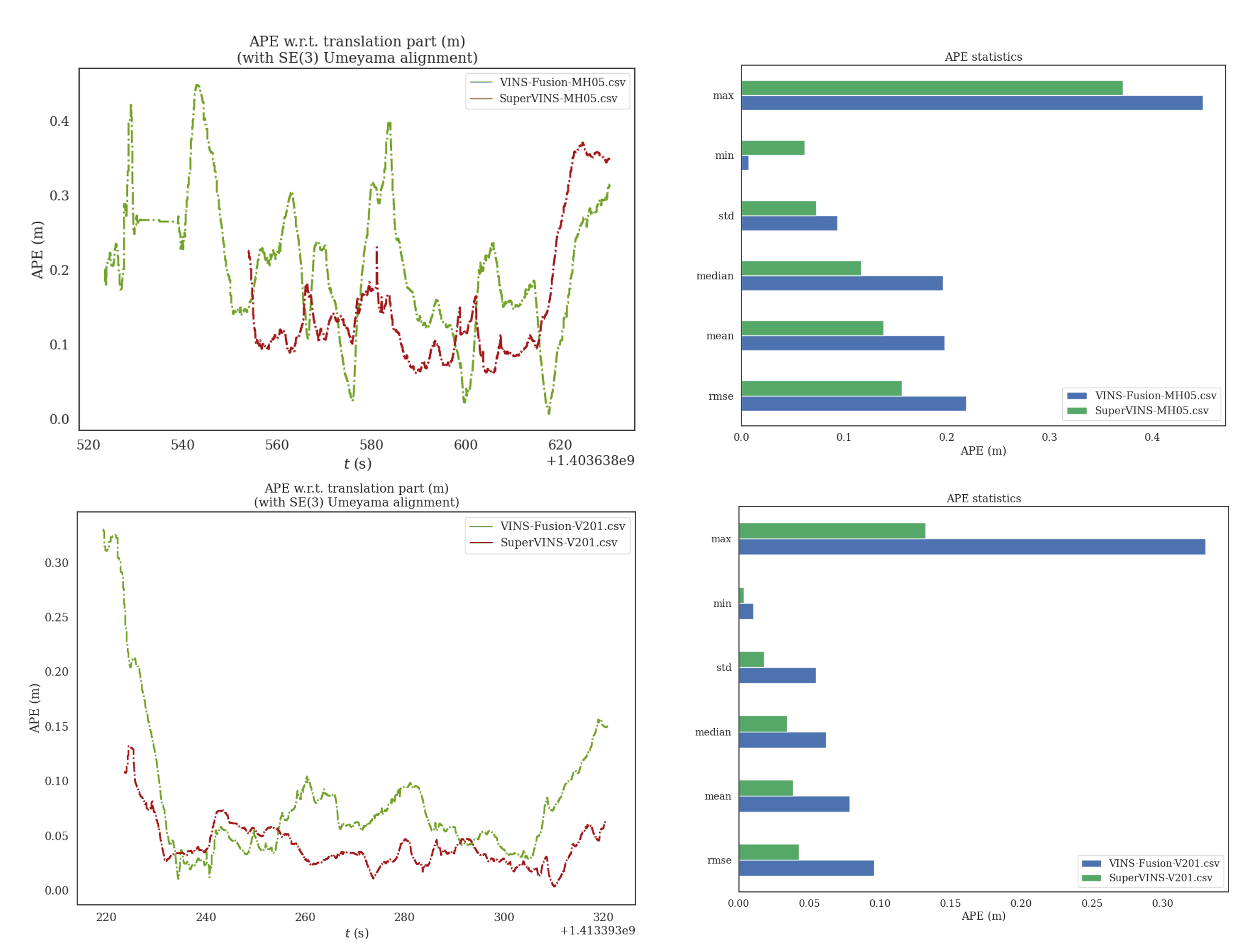}
    \caption{Error timing comparison between SuperVINS and VINS-Fusion in
MH05 and V201 sequences.}
    \label{compare_error1}
\end{figure}

\begin{figure}[H]
    \centering
    \includegraphics[width=0.5\linewidth]{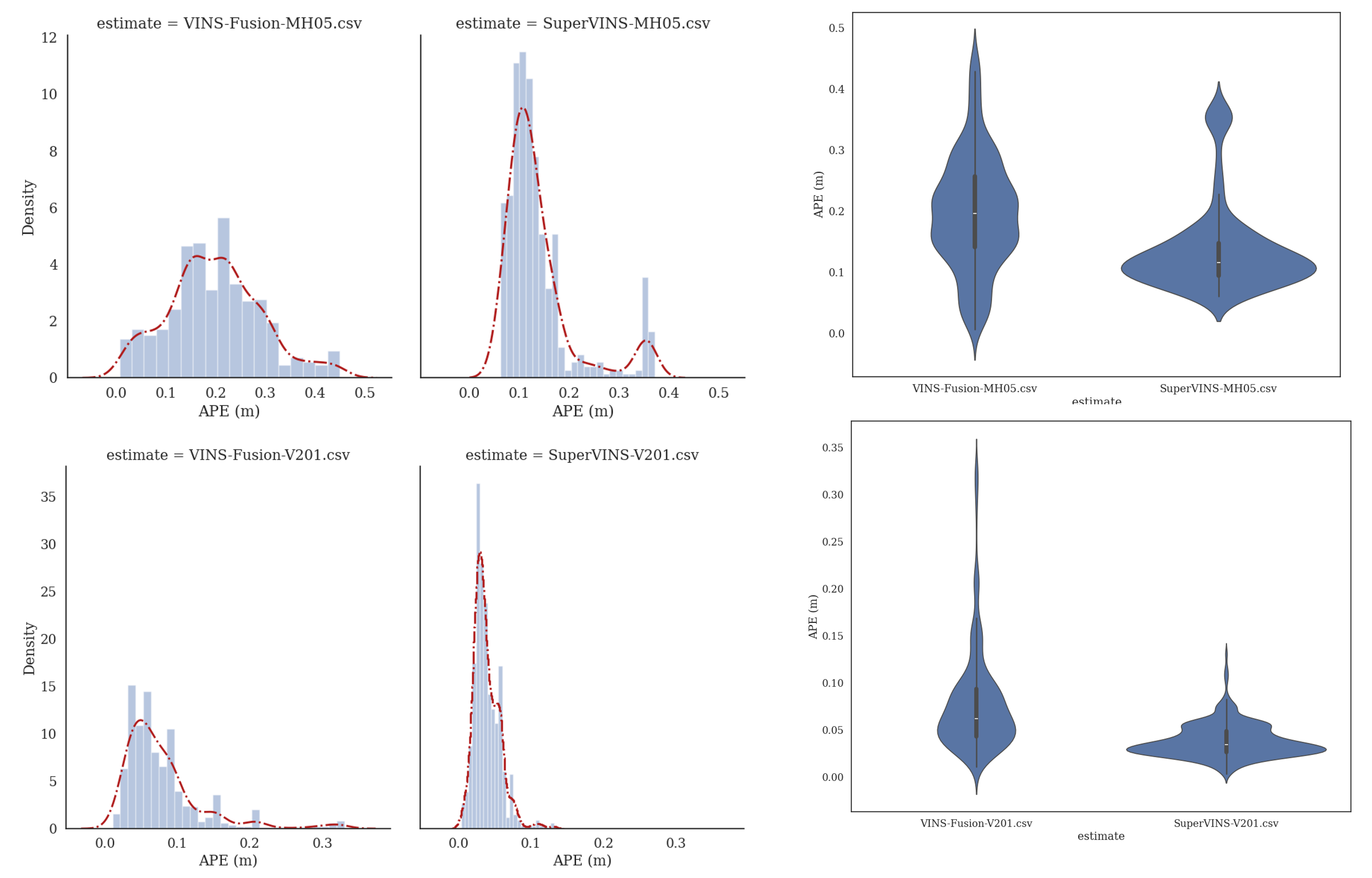}
    \caption{ Comparison of error distribution between SuperVINS and VINSFusion in MH05 and V201 sequences.}
    \label{compare_error2}
\end{figure}

In the comparison between SuperVINS and VINS-Fusion, both algorithms were equipped with loop closure detection modules to ensure fairness. Values representing relatively excellent experimental results are highlighted in bold. Among the 11 tested sequences, SuperVINS demonstrated advantages in 6, particularly in the challenging sequences MH05 and V203. Moreover, the performance of the improved SuperVINS algorithm in the remaining 5 sequences did not exhibit a significant decline. Analyzing the data presented in TABLE \ref{vins-fusion_SuperVINS_compare}, we can draw the following conclusions:
\begin{enumerate}{}{}
    \item{In the MH02, MH03, and MH04 sequences, characterized by rich textures, normal camera motion speeds, and strong visual features, the localization accuracy of SuperVINS is slightly lower than that of VINS-Fusion. This can be attributed to the fact that in such conventional environments, traditional geometric features are adequate, thereby limiting the advantage of deep learning-based feature point extraction. However, in scenarios involving rapid motion and low lighting, the superiority of the deep learning-based feature extraction approach becomes evident. For instance, in the extremely challenging V203 sequence, the Absolute Trajectory Error (ATE) improved by 12.3\%. In the particularly difficult MH05 sequence, the root mean square error (RMSE) of the absolute position error (ATE) was 0.158304, representing a 39.6\% improvement over the ATE of VINS-Fusion. This enhancement in accuracy is significant.}
    \item{Compared to VINS-Fusion, SuperVINS demonstrated advantages in 6 out of the 11 sequences, while the remaining five sequences showed no significant decline in accuracy. This indicates that the SuperVINS algorithm not only maintains real-time performance but also enhances localization accuracy. For instance, in the MH05 and V201 sequences, the absolute position error  exhibited substantial improvements across various metrics, including maximum, minimum, average, median values, and the residual sum of squares, as illustrated in Fig.\ref{compare_error1}.}
    \item {This section visualizes the errors in two relatively challenging sequences, MH05 and V201, with the results presented in Fig.\ref{compare_error1} and Fig.\ref{compare_error2}. A clear distinction is evident between VINS-Fusion and SuperVINS. The red line in the figures indicates the error distribution of the SuperVINS trajectory, demonstrating a significant reduction in trajectory error across both challenging sequences. Additionally, a statistical analysis of the errors reveals substantial decreases in the maximum, minimum, and median values. Therefore, we conclude that the improved algorithm exhibits superior performance in the selected challenging sequences.}
    \item {This section employs a Gaussian function to model the error distribution. The results presented in Fig.\ref{compare_error2} indicate that both VINS-Fusion and SuperVINS errors conform to a Gaussian distribution. Notably, the improved Gaussian fitting curve for SuperVINS shifts to the left, signifying a reduction in overall error. Furthermore, an analysis of the Gaussian curves reveals intriguing differences in error distribution: the curve for SuperVINS is narrower, while that for VINS-Fusion is wider. This suggests that the stability of the error distribution has been enhanced to a considerable extent.}
\end{enumerate}
\subsection{Real-time and Robustness Testing of SuperVINS}
In the SuperVINS system, the front-end feature point extraction and matching processes leverage deep learning techniques. While deep learning features offer advantages over traditional geometric features, they also demand greater computational resources. Ensuring real-time performance is thus a critical consideration for the system. Consequently, this paper presents a real-time test of the system's performance.
\begin{figure}[H]
    \centering
    \includegraphics[width=0.8\linewidth]{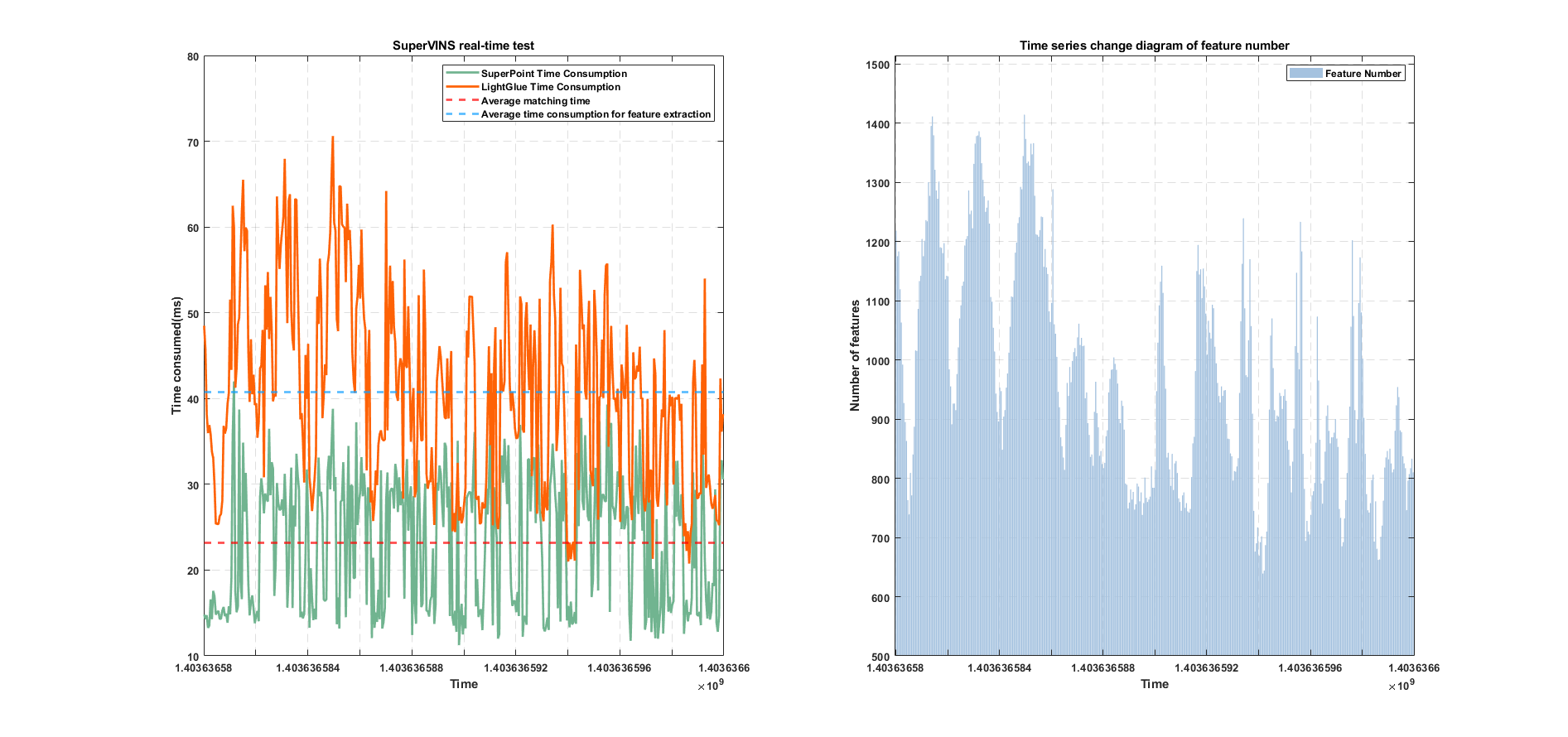}
    \caption{Schematic diagram of time consumption for deep learning feature extraction and matching.}
    \label{realtime}
\end{figure}
This paper tests the MH01 sequence data set, and the selected timestamp is 1403636580.01355-1403636599.96355. The time of this timestamp is nearly 20s. According to the experimental results, it can be obtained that the feature point extraction is 43fps, the feature matching is 24.5fps, and the average number of feature points is 965. This basically meets the real-time requirements.

\begin{figure}[H]
    \centering
    \includegraphics[width=0.8\linewidth]{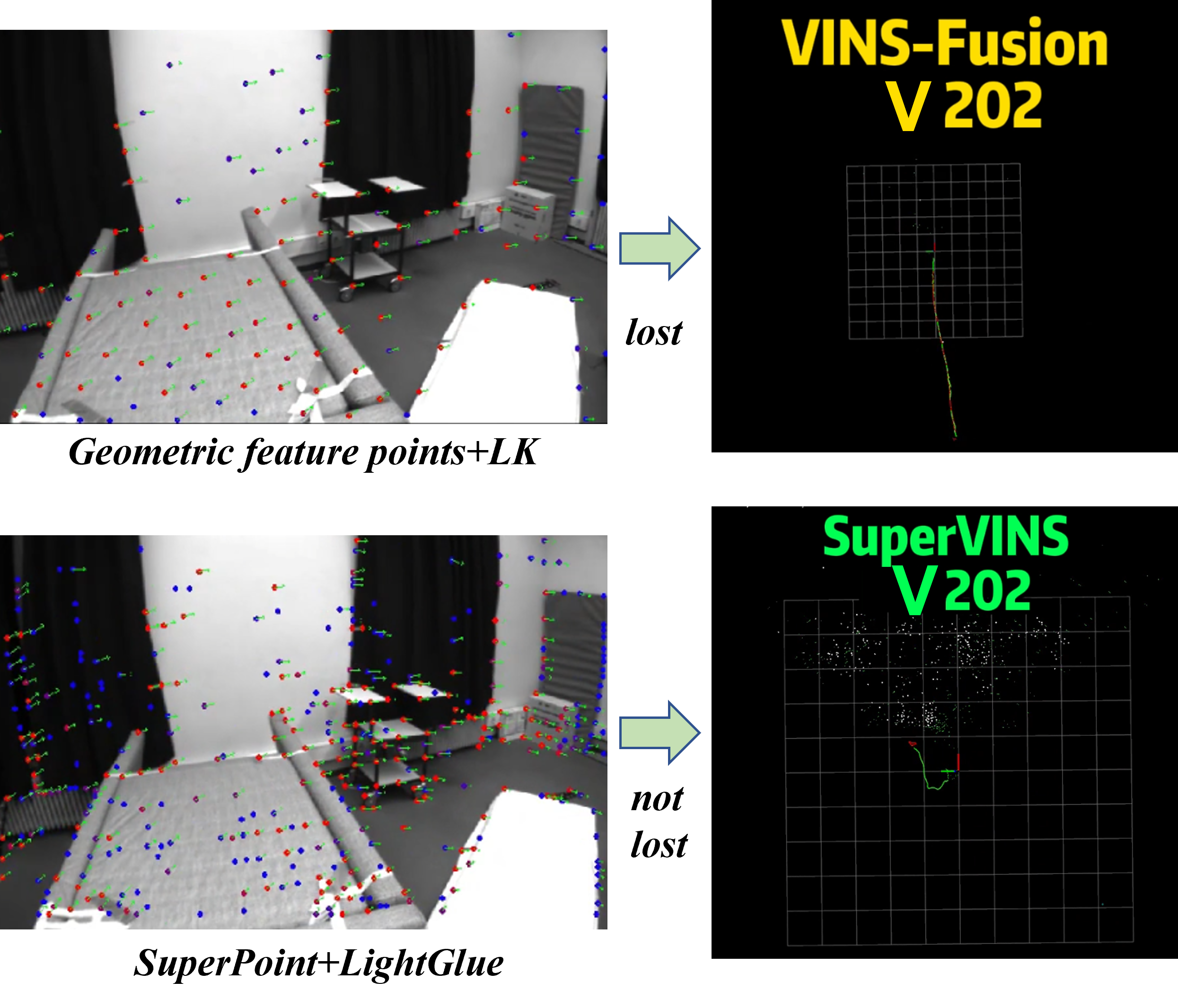}
    \caption{Robustness comparison between VINS-Fusion and SuperVINS.}
    \label{robustness}
\end{figure}

While VINS-Fusion is a well-regarded classic algorithm, its robustness can be further improved. The SuperVINS system, which employs highly robust feature points and matching methods, demonstrates superior stability compared to VINS-Fusion. As illustrated in Fig. \ref{robustness}, both systems yield different results when processing the same V202 sequence dataset. VINS-Fusion ultimately experiences localization failures due to the sparse extraction of feature points and rapid camera motion, which causes image blur. In contrast, the SuperVINS system maintains effective localization throughout the V202 sequence, ensuring reliable localization and mapping. This comparison underscores the enhanced robustness of SuperVINS over the classic VINS-Fusion algorithm.

\section{Conclusion}
This paper presents SuperVINS, a real-time visual-inertial SLAM framework built upon deep learning neural networks. We introduce a matching enhancement strategy that employs a variable threshold based on the RANSAC algorithm. Additionally, we retrain the deep learning-based Bag of Words and integrate it into the loop detection module of the visual-inertial SLAM system. Experimental validation on a public dataset demonstrates considerable improvements in localization accuracy, real-time performance, and robustness of the system.

While the proposed method has achieved significant progress, there remains considerable scope for enhancement. For instance, the current optimization of the matching strategy overlooks environmental adaptability, necessitating a flexible approach for threshold selection tailored to specific scenarios. Additionally, future research could investigate lightweight implementations of point-line fusion features leveraging deep learning.

{

}

\end{document}